\documentclass[10pt,twocolumn,letterpaper]{article}

\usepackage[pagenumbers]{cvpr} 

\definecolor{cvprblue}{rgb}{0.21,0.49,0.74}
\usepackage[pagebackref,breaklinks,colorlinks,allcolors=cvprblue]{hyperref}

\def\paperID{*****} 
\def\confName{CVPR}
\def\confYear{2026}

\title{SignDATA: Data Pipeline for Sign Language Translation}

\author{Kuanwei Chen\thanks{Corresponding author.}\\
Computer Science and Information Engineering\\
National Central University, Zhongli, Taiwan\\
{\tt\small elaping5691@gmail.com}
\and
Tingyi Lin\\
Electrical Engineering\\
National Changhua University Of Education, Changhua, Taiwan\\
{\tt\small brian104120@gmail.com}
}

\begin{document}
\maketitle
\begin{abstract}
Sign-language datasets are difficult to preprocess consistently because they vary in
annotation schema, clip timing, signer framing, and privacy constraints. Existing work
usually reports downstream models, while the preprocessing pipeline that converts raw
video into training-ready pose or video artifacts remains fragmented, backend-specific,
and weakly documented. We present SignDATA, a config-driven preprocessing toolkit that
standardizes heterogeneous sign-language corpora into comparable outputs for learning.
The system supports two end-to-end recipes: a pose recipe that performs acquisition,
manifesting, person localization, clipping, cropping, landmark
extraction, normalization, and WebDataset export, and a video recipe that replaces pose
extraction with signer-cropped video packaging. SignDATA exposes
interchangeable MediaPipe and MMPose backends behind a common interface, typed job
schemas, experiment-level overrides, and per-stage checkpointing with config- and
manifest-aware hashes. We validate the toolkit through a research-oriented evaluation
design centered on backend comparison, preprocessing ablations, and privacy-aware video
generation on datasets. Our contribution is a reproducible preprocessing layer for sign-language
research that makes extractor choice, normalization policy, and privacy tradeoffs explicit,
configurable, and empirically comparable.
Code is available at \url{https://github.com/balaboom123/signdata-slt}.
\end{abstract}
    
\section{Introduction}

Sign language translation (SLT) has advanced rapidly in recent years, driven by improvements in sequence modeling, 
the release of larger public datasets, and growing interest in end-to-end translation from 
sign videos to spoken-language text. However, progress in SLT remains strongly 
constrained by data availability and by the practical difficulty of converting raw sign language 
videos into research-ready training inputs.
Public sign language datasets differ substantially in scale, collection conditions, annotation structure,s
and usage constraints. For example, How2Sign is a large-scale multimodal dataset collected in a controlled studio environment 
with more than 80 hours of instructional American Sign Language (ASL) data~\cite{how2sign2021}, whereas YouTube-ASL is a much 
larger open-domain ASL-English parallel corpus with approximately 984 hours of captioned video and more than 2{,}
500 unique signers~\cite{youtubeasl2023}. These differences make dataset-specific preprocessing a recurring and non-trivial part of 
SLT research.

At the same time, recent work has made clear that preprocessing is not only an implementation detail, but an important part of the 
research pipeline itself. As SLT systems scale toward larger and more diverse video collections, researchers must address 
not only heterogeneous video formats and annotation sources, but privacy concerns associated with biometric information 
contained in raw sign language videos~\cite{privacyslt2024}. This motivates the need for preprocessing frameworks that are reproducible, 
extensible, and mindful of practical research constraints, rather than writing one-time code that cannot be reused.

In this work, we present \textbf{SignDATA}, an open-source, config-driven preprocessing framework for sign language research. 
SignDATA currently supports both How2Sign and YouTube-ASL~\cite{how2sign2021,youtubeasl2023}, and provides two complementary output modes:
landmark-based pose representations, which have been widely used as an alternative representation to raw RGB videos in prior sign 
language research, and clipped video segments. Recent work, however, highlights that detailed facial landmarks still carry 
biometric risks and therefore should not be assumed to provide meaningful privacy protection on their own~\cite{privacyslt2024}. 
By supporting both pose-based and video-based outputs within the same preprocessing framework, SignDATA allows researchers to choose 
representations according to their task requirements, computational constraints, and privacy considerations. 
The framework further supports multiple extraction backends, such as but not limited to MediaPipe
Holistic~\cite{mediapipe_holistic_guide} and MMPose~\cite{mmpose_github}.
Instead of hard-coding a single dataset or preprocessing recipe, SignDATA organizes preprocessing through YAML-based configuration, 
modular pipeline stages, multi-worker parallel processing, and registry-based extension points for datasets, processors, and extractors.

This design targets several practical needs in current SLT research. First, it reduces repeated engineering effort when preparing 
different datasets under a unified workflow. Second, it lowers the barrier for new researchers by exposing preprocessing behavior 
through documented configuration files rather than deeply entangled code paths. Third, it improves maintainability and community 
contribution by separating dataset logic, processing steps, and extraction backends through a modular registry-based architecture. 
Finally, because the framework is fully open source and explicitly distinguishes repository licensing from the licenses of external 
datasets and upstream tools, it is suitable as a
reusable infrastructure layer for broader sign language AI research.

The contribution of this work is not a new SLT model, but a reusable preprocessing architecture for sign language research. 
We provide a unified pipeline that spans dataset handling, manifest construction, landmark extraction or video clipping, 
post-processing such as landmark reduction and normalization, and export to training-ready shard formats such as WebDataset. 
Through this framework, we aim to reduce repetitive engineering effort in sign language data preparation while improving reproducibility,
extensibility, and accessibility for both new and experienced researchers.
\section{Related Work}

\subsection{Sign Language Datasets and Data Scaling}

A major challenge in sign language translation (SLT) research is the limited availability of large-scale, 
high-quality parallel data. Earlier SLT studies were often conducted on relatively small and domain-constrained datasets, 
which made it difficult to assess model robustness under realistic variation in signers, recording conditions, and language use. 
More recent datasets have significantly expanded the scope of SLT research by providing richer annotations and larger video collections.

How2Sign~\cite{how2sign2021} is one of the influential datasets in this direction. It provides a large-scale multimodal corpus
of continuous American Sign Language (ASL) collected in a controlled studio environment, with more than 80 hours of aligned sign videos,
speech, English transcripts, and additional modalities. In contrast, YouTube-ASL~\cite{youtubeasl2023} shifts the focus toward scale
and diversity by introducing an open-domain ASL-English parallel corpus mined from YouTube, featuring approximately 984 hours of video
and more than 2,500 unique signers. OpenASL~\cite{openASL2022} similarly pursues open-domain coverage, providing a large-scale
ASL-English corpus sourced from online video with automatically derived alignments.
Beyond ASL, PHOENIX-2014T~\cite{phoenix2014t} has served as the standard benchmark for German Sign Language (DGS) translation,
providing sentence-level gloss and text annotations from broadcast weather forecasts. CSL-Daily~\cite{csldaily2021} extends SLT
research to Chinese Sign Language with approximately 20,000 annotated sentences covering everyday topics.
BOBSL~\cite{bobsl2021} provides approximately 1,400 hours of British Sign Language video sourced from broadcast television,
with weakly supervised subtitle alignments. At the isolated-sign level, WLASL~\cite{wlasl2020} offers over 21,000 video instances
spanning 2,000 ASL signs, while ASL-Citizen~\cite{aslcitizen2023} contributes roughly 83,000 community-sourced videos covering
2,731 signs collected from Deaf signers.
Together, these datasets illustrate an important challenge for SLT preprocessing: different datasets may differ not only
in scale, but also in acquisition assumptions, annotation format, visual quality, and licensing conditions.

For this reason, preprocessing in SLT is not merely a fixed preparation step before training. Instead, it must adapt to heterogeneous 
datasets while preserving reproducibility across experiments. This motivates preprocessing systems that can accommodate both controlled 
datasets such as How2Sign and large open-domain datasets such as YouTube-ASL under a unified and reusable framework.

\subsection{Video-Based Sign Language Translation}

A major line of SLT research operates directly on RGB video and learns a mapping from spatio-temporal visual input to 
spoken-language text. Sign Language Transformers~\cite{camgoz2020sign} established an important transformer-based 
baseline for joint continuous sign language recognition and translation, showing that recognition and translation can be learned 
together in an end-to-end manner. This work helped establish transformer architectures as a strong foundation for SLT.

Subsequent work further explored how translation quality depends on the role of intermediate supervision and representation design. 
STMC-Transformer~\cite{yin2020better} showed that transformer-based SLT can outperform earlier recurrent approaches and argued that 
ground-truth gloss translation should not be assumed to define the upper bound of SLT performance, as their video-to-text model 
surpassed translation from ground-truth glosses. More recent methods have continued to improve video-based SLT through transfer learning,
 unified modeling, and visual-language pretraining~\cite{chen2022simple,zhang2023sltunet,zhou2023glossfree}.

Taken together, these studies show that raw video remains a central 
representation in SLT. At the same time, they also imply that building stable and reproducible 
video-based SLT systems requires more than model design alone. Differences in data sources, visual conditions, and preprocessing 
assumptions can all affect how video is prepared and consumed by downstream models. This makes video preprocessing an important 
systems-level concern rather than a purely incidental implementation detail.

\subsection{Pose- and Keypoint-Based Representations and Preprocessing}

In parallel with RGB-based approaches, many studies have explored pose- or keypoint-based sign representations. 
These methods typically transform video into structured landmarks for the body, hands, and face, then model the resulting sequences 
with recurrent, transformer-based, or graph-based architectures. Pose-based methods are attractive because they can reduce background 
variation, simplify the input space, and provide a compact representation of signing motion.

Pose-based work has shown that preprocessing choices are highly consequential rather than incidental,
as decisions such as normalization strategy, keypoint selection, and handling of missing detections can directly affect
downstream recognition and translation performance~\cite{youtubeasl2023,camgoz2020sign,chen2022simple}.

These works are closely related to our motivation. However, most prior studies present preprocessing as a paper-specific 
procedure attached to a particular task, dataset, or model. For instance, the baseline pipeline introduced with YouTube-ASL 
uses MediaPipe Holistic landmarks, reduces the original 532 landmarks to 85 selected points, and normalizes them to a unit 
bounding box before translation modeling~\cite{youtubeasl2023}. In contrast, our work focuses on preprocessing itself as reusable 
research infrastructure. This includes dataset handling, manifest generation, landmark extraction or video clipping, normalization, 
and export into training-ready formats within a unified pipeline.

From an implementation perspective, pose-based pipelines also depend heavily on the choice of extraction backend.
OpenPose~\cite{openpose2017} was among the first systems to demonstrate real-time multi-person 2D pose estimation using
Part Affinity Fields, establishing a widely adopted baseline for body, hand, and face keypoint extraction.
BlazePose~\cite{blazepose2020} later introduced an on-device, single-person pose tracker optimized for mobile and real-time
use cases, forming the foundation of the MediaPipe Holistic pipeline~\cite{mediapipe_holistic_guide}, which provides
integrated face, hand, and body landmark extraction at real-time performance.
MMPose~\cite{mmpose_github} provides a complementary open-source pose estimation ecosystem with strong practical
performance and flexible deployment. These tools make pose-based SLT increasingly accessible, but they also reinforce the need for
frameworks that can organize extractor-specific decisions in a modular and reproducible way.

\section{Problem Setup and Design Goals}
\label{sec:problem}

\subsection{Problem Setup}

We study preprocessing as a systems problem. The input is a sign-language dataset
consisting of raw videos together with dataset-specific metadata such as transcript
files, alignment CSVs, split labels, signer identifiers, timing boundaries, or auxiliary
bounding boxes. These inputs are heterogeneous by construction: YouTube-ASL uses web
acquisition and transcript JSON parsing, How2Sign expects pre-downloaded videos plus an
existing alignment CSV, and OpenASL~\cite{openASL2022} is implemented through a TSV-based adapter with
optional bounding-box metadata.

The output is one of two standardized artifact types. In \textbf{pose mode}, the system
emits frame-wise landmark tensors that can be normalized, reduced to preset keypoint
subsets, flattened, and exported as WebDataset shards for model training. In
\textbf{video mode}, the system emits processed video clips, optionally cropped and
obfuscated, and packages them with aligned metadata in the same export format. In both
cases, the preprocessing run is defined by a job configuration rather than by manual
scripting.

\subsection{Design Goals}

The implementation is guided by five design goals.
\begin{itemize}
    \item \textbf{Modularity.} Dataset acquisition, manifest creation, detection,
          clipping, cropping, extraction, normalization, obfuscation, and packaging
          should exist as independent stages.
    \item \textbf{Reproducibility.} Runs should be represented by typed configuration
          objects, explicit overrides, and checkpoint markers rather than by hidden
          state in notebooks or scripts.
    \item \textbf{Backend interchangeability.} Extractor backends should be swappable at
          the configuration level without rewriting downstream stages.
    \item \textbf{Dataset portability.} Dataset adapters should terminate in a canonical
          manifest schema so that shared processors can operate across corpora.
    \item \textbf{Scalability.} The pipeline should support multi-worker parallel processing
          and export to optimized sharded formats (e.g., WebDataset~\cite{webdataset}) to mitigate I/O and storage
          bottlenecks when scaling to massive open-domain collections.
\end{itemize}

These goals constrain the scope of the paper. We do not claim a new pose estimator or a
universal landmark ontology across toolkits. Instead, we claim a preprocessing toolkit
that makes heterogeneous dataset preparation explicit, repeatable, and experimentally
comparable.

\section{System Overview}
\label{sec:system}

\begin{figure*}[t]
    \centering
    \includegraphics[width=0.96\linewidth]{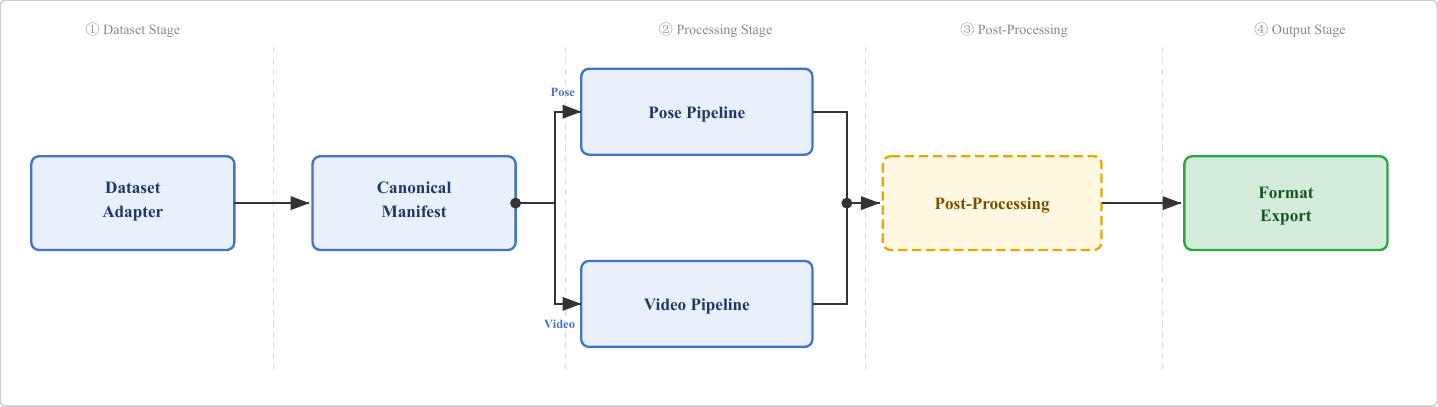}
    \caption{End-to-end SignDATA pipeline. Dataset adapters first acquire or validate
             raw data and produce a canonical manifest. The pose pipeline extracts
             skeletal landmarks and applies optional post-processing normalization;
             the video pipeline crops and segments raw footage. Both pipelines
             export processed outputs to configurable sharded formats for downstream
             training.}
    \label{fig:pipeline}
\end{figure*}

Figure~\ref{fig:pipeline} shows the end-to-end pipeline.
SignDATA follows a config-driven execution model. At startup, the system loads dataset,
processing, and  output components into a set of registries so they can be selected by
name at runtime. A job or experiment is declared in a YAML configuration file, which is
validated and used to drive execution. All pipelines follow an ordered stage sequence
described in Section~\ref{sec:method}.
The framework currently supports both \textbf{pose} and \textbf{video} pipelines, with
person detection handled within the processing step rather than exposed as a separate
top-level stage, and outputs packaged into sharded formats such as WebDataset.

Shared state is passed through a context object that carries the validated configuration,
the active dataset adapter, run-scoped output directories, and per-stage execution
metadata. Each run is assigned deterministic output paths so that artifacts from
different runs remain isolated. Stages communicate through a small fixed set of agreed
artifact locations and in-memory manifest state, giving the system stable handoff points
between acquisition, processing, post-processing, and packaging without requiring
downstream components to know dataset-specific details.

The system supports reproducibility through typed and mergeable YAML configuration,
deterministic run-scoped artifact paths, explicit stage ordering, and experiment files
that execute ordered job lists with per-job configuration overrides. A finer-grained
checkpoint layer is partially implemented, but it is not yet part of the stable
execution contract in the current release.
\section{Pipeline Components}
\label{sec:method}

\subsection{Dataset Stage}

Dataset adapters isolate source-specific acquisition logic and convert each source into a
canonical manifest schema shared across the pipeline. The schema requires only sample and
video identifiers; timing, label, spatial, split, and metadata fields are optional. A fixed
set of known column aliases is normalized on load so that downstream stages consume a
uniform representation.

Acquisition logic varies by source: some adapters download both video and transcript
data, as in YouTube-ASL-style workflows; others assume locally prepared dataset releases
and read provided annotations from a release directory structure, as in How2Sign-style
workflows. Other adapters may read a provided annotation file and optionally download
missing video files. Segments are filtered by rule-based criteria---presence of required
text and timing fields, text normalization, and configurable duration bounds---before
being written to the canonical manifest consumed by all subsequent stages.

\subsection{Processing Stage}

The processing layer offers two active modes.

\textbf{Pose pipeline.} The pose mode reads temporally aligned segments from the manifest,
samples frames at a configurable rate, and optionally runs person detection to locate the signer.
Person detection can be performed using a detector such as YOLO~\cite{yolov8_ultralytics}
or MMDetection~\cite{mmdetection}, and the resulting
bounding boxes are forwarded to a selectable pose backend. The pipeline currently supports
multiple pose frameworks as interchangeable options, including MediaPipe Holistic and
MMPose, each producing per-frame keypoint arrays in a shared structured format, though the
number of keypoints is backend-specific.

\textbf{Video pipeline.} The video mode uses a two-pass decoding workflow: the first pass
samples frames for detection and computes a bounding region around the signer; the second
pass clips and crops the source video to that region, with optional padding and resizing
applied. Multi-person clips and clips with no detection are skipped and not carried
forward. This design is consistent with the practical filtering requirements commonly
encountered in open-domain web video corpora such as YouTube-ASL~\cite{youtubeasl2023}.

\subsection{Post-Processing Stage}

Post-processing is an optional transformation stage applied between processing and
output. It accepts processed outputs and applies any configured transformations to
condition the data before it is handed to the output stage, allowing additional
refinement steps to be inserted as needed. This design provides flexibility for
representation-specific operations such as normalization without changing the higher-level
pipeline structure. When post-processing is not configured or is bypassed, outputs are
carried forward unchanged.

\subsection{Output Stage}

Processed outputs are written into WebDataset~\cite{webdataset} sharded tar archives suitable for direct
consumption by training pipelines. Pose jobs package normalized keypoint arrays after
post-processing, or raw arrays if post-processing is not applied. Video jobs package
cropped video clips. Each sample additionally stores caption text when available and a
metadata record containing sample and video identifiers, timing boundaries, and the
processor used. Shards are organized under run-scoped output directories, so outputs from
different runs remain isolated without manual intervention.

\section{Discussion}
\label{sec:discussion}

\subsection{Design Tradeoffs}

The SignDATA codebase supports treating preprocessing as a structured design space
rather than a fixed recipe~\cite{sculley2015hidden}. Choices at the dataset-adapter, sampling, detection,
pose-backend, normalization, and output levels each alter both the information retained
and the failure modes introduced. MediaPipe and MMPose should therefore be understood
as different representation choices rather than interchangeable extractors: they differ
in landmark semantics, keypoint coverage, and the conditions under which they fail.
Similarly, pose-only and cropped-video outputs represent a trade-off among privacy,
storage, compute, and retained visual information~\cite{privacyslt2024} rather than a simple quality ranking.
Neither backend nor output mode is universally preferable; the appropriate choice is
determined by dataset conditions, available compute, and the requirements of the
downstream task.

\subsection{Goal Retrospective}

Revisiting the design goals stated in Section~\ref{sec:problem}: \textit{Modularity} is
met through the stage-based architecture, in which acquisition, detection, extraction,
normalization, and packaging exist as independent, composable units.
\textit{Reproducibility} is substantially achieved via typed YAML configuration and
deterministic run-scoped artifact paths~\cite{sculley2015hidden, pineau2021reproducibility},
though the checkpoint layer remains partially integrated.
\textit{Backend interchangeability} is realized at the configuration level for
different processors, though cross-backend landmark semantics remain non-equivalent.
\textit{Dataset portability} is met through the registry-based adapter architecture and
the canonical manifest schema~\cite{hutchinson2021accountability, gebru2021datasheets}.
\textit{Scalability} is addressed through multi-worker parallel processing and WebDataset
shard export~\cite{webdataset}; the remaining gap is the absence of runtime
stage-skipping on restart.

\subsection{Limitations}

The current implementation carries several explicit boundaries.
\textit{Semantic and scope limitations:} cross-backend landmark semantics are not
equivalent even when both outputs are reduced to the same dimensionality through keypoint
presets~\cite{mediapipe_holistic_guide, mmpose_github}, so users should commit to a single
backend before designing downstream model inputs, as switching backends mid-project
requires re-extracting all data; and the active processing path assumes a single signer
per segment and drops multi-person clips, meaning open-domain corpora should be
pre-filtered or paired with an upstream multi-person detector to limit clip
loss~\cite{youtubeasl2023, bobsl2021}.
\textit{Implementation gaps:} WebDataset metadata is intentionally minimal and does not
preserve every manifest field~\cite{webdataset}, so users requiring full manifest fields
should join against the manifest CSV post-hoc; and checkpoint utilities exist but are not
yet integrated into runtime stage skipping, meaning restarted runs may reprocess
already-completed shards and users should verify shard counts manually before continuing.

These are scoping decisions rather than fundamental dead ends, and they define clear
extension points for future work.

\section{Ethics}

This work introduces \textbf{SignDATA}, an open-source preprocessing framework and repository for sign language research.
Our primary goal is to improve reproducibility, extensibility, and accessibility in sign language data preparation workflows,
especially for research and educational use. We do not present this framework as a tool for signer identification, biometric
profiling, surveillance, or other identity-centric applications.

Unlike general video datasets in which subjects are incidental to the content, sign language videos are intentionally
subject-centric: the signer's face and body motion constitute the communicative signal itself, making it structurally
difficult to separate informative content from biometric data. Sign language datasets therefore consist of raw videos
containing faces, upper-body appearance, and motion patterns that may reveal biometric information. This concern is
compounded at web scale: collection from open-domain platforms such as YouTube-ASL~\cite{youtubeasl2023} and
OpenASL~\cite{openASL2022} raises additional questions around informed consent and the representational implications
of aggregating signing behavior across diverse signer communities~\cite{jo2020lessons}.
Recent work has explicitly foregrounded these risks~\cite{privacyslt2024}, arguing that scaling SLT should account
for biometric information in sign videos and cautioning that sufficiently detailed facial landmarks may still function
as biometric identifiers rather than meaningful anonymization. The SignDATA repository supports both clipped video
outputs and landmark-based pose representations; these pose-based outputs should therefore be understood as a
representational option for research, not a guarantee of privacy preservation. Dataset documentation frameworks
such as Datasheets for Datasets~\cite{gebru2021datasheets,hutchinson2021accountability} further underscore that
transparency in data collection and intended use is central to responsible dataset development, and we view
preprocessing as a stage where such responsible practices should be actively considered.

The repository is released under the MIT License for its code and documentation, and its README explicitly distinguishes this
software license from the licenses governing external datasets, research artifacts, and upstream repositories used with the
framework~\cite{how2sign2021,youtubeasl2023,mediapipe_holistic_guide,mmpose_github}. Users remain responsible for ensuring that
data acquisition, preprocessing, storage, redistribution, and downstream use comply with the original licenses and usage terms
of the relevant datasets and dependencies.

We therefore encourage the use of this repository in ways that respect signer privacy, data governance requirements, and the
broader interests of sign language communities. A central concern in this regard is ensuring that sign language technology
research is developed in meaningful partnership with Deaf communities, rather than treating signers solely as data
sources~\cite{bragg2019sign}. Furthermore, users should remain aware that public SLT datasets may reflect
important representational limitations and demographic biases. For example, How2Sign explicitly notes limited diversity in
race/ethnicity and skin tone~\cite{how2sign2021}, while YouTube-ASL highlights variation across dialect and signer skill
level~\cite{youtubeasl2023}. Recent work has also cautioned that SLT models may underperform for underrepresented demographic
groups and may fail to adequately capture regional variation~\cite{privacyslt2024}. While SignDATA standardizes the preprocessing
pipeline, models trained on the resulting outputs will still reflect the representational limitations of the underlying data.

\section{Conclusion}
\label{sec:conclusion}

This paper presents SignDATA, a reproducible, config-driven preprocessing toolkit for
sign-language datasets. By standardizing heterogeneous corpora into comparable pose
tensors or clipped video artifacts through dataset adapters, recipe-driven execution,
interchangeable extractor backends, and explicit normalization policies, SignDATA turns
backend choice, clipping policy, normalization, and privacy trade-offs into explicit,
reproducible experimental variables --- rather than hidden implementation choices.

Our backend comparison reveals that MediaPipe and MMPose are distinct representation
choices, not interchangeable extractors --- differing
meaningfully in throughput, landmark coverage, and failure modes across controlled and
open-domain conditions. The preprocessing ablations further demonstrate that individual
decisions --- signer-region cropping, visibility masking, and depth coordinate retention
--- produce measurable, attributable differences in usable-clip rate and storage
footprint.

Several directions remain open. Runtime checkpoint integration would eliminate redundant
reprocessing on interrupted runs. Planned privacy-preserving video obfuscation will let
researchers produce signer-anonymized outputs without altering the pipeline structure.
Cross-backend landmark ontology normalization would make pose sequences extracted by
different backends directly comparable, reducing the cost of extractor switching.
Finally, expanding dataset adapter coverage beyond the ASL-centric corpora evaluated
here would broaden SignDATA's utility to a wider range of signed languages.

{
    \small
    \bibliographystyle{ieeenat_fullname}
    \bibliography{main}
}


\end{document}